\documentclass[12pt]{amsart}
\usepackage[utf8]{inputenc}
\usepackage[utf8]{inputenc}
\usepackage[T1]{fontenc}
\usepackage[all]{xy}
\usepackage{lipsum}
\usepackage{url}
\usepackage{tikz}
\usepackage{stackrel}
\usepackage{color}
\def\red{}
\def\blue{}
\usepackage{amsmath,amsthm,amssymb}

\usepackage{xcolor}

\usepackage{graphicx}

\newtheorem{theorem}{Theorem}[section]

\newtheorem{example}[theorem]{Example}

\newtheorem{proposition}[theorem]{Proposition}
\theoremstyle{definition}
\newtheorem{definition}[theorem]{Definition}

\newtheorem{remark}[theorem]{Remark}
\newtheorem{corollary}[theorem]{Corollary}
\numberwithin{equation}{section}

\begin{document}

\renewcommand{\bf}{\bfseries}
\renewcommand{\sc}{\scshape}

\title[Parametrised motion planning algorithms]%
{Parametrised collision-free optimal motion planning algorithms in Euclidean spaces}

\author{Cesar A. Ipanaque Zapata}
\address{Departamento de Matem\'{a}tica, Universidade de S\~{a}o Paulo Instituto de Matem\'{a}tica e  Estatística -IME/USP, R. do Matão, 1010 - Butantã, CEP:
05508-090 - S\~{a}o Paulo, Brasil}
\email{cesarzapata@usp.br}
\thanks{The first author would like to thank grant\#2016/18714-8 and grant\#2022/03270-8, S\~{a}o Paulo Research Foundation (FAPESP) for financial support.}

\author{Jes\'{u}s Gonz\'{a}lez}
\address{Departamento de Matem\'{a}ticas, Centro de Investigaci\'{o}n y de Estudios Avanzados del I.P.N.
Av.~Instituto Polit\'{e}cnico Nacional n\'{u}mero 2508,
San Pedro Zacatenco, M\'exico City 07000, M\'{e}xico}
\email{jesus@math.cinvestav.mx}



\begin{abstract} We \red{describe} parametrised motion planning algorithms \red{for} systems controlling objects \red{represented by points} that move \red{without collisions} in \red{an even dimensional} Euclidean space and in the presence of \red{up to three} obstacles with \red{\emph{a priori} unknown} positions. Our algorithms are optimal in \red{the sense that the parametrised} local planners \red{have minimal posible size.} 
\end{abstract}
\maketitle

\section{Introduction \red{and main results}}\label{secintro}
The design of explicit motion planners that are reasonably close to optimal is one of the challenges \red{in} modern robotics (see for \red{instance} Latombe \cite{latombe2012robot} and LaValle \cite{lavalle2006planning}). \red{As an answer to such a need,} the concept of parametrised topological complexity \red{has recently been} introduced in \cite{cohen-farber-weinberger} by Cohen, Farber and Weinberger \red{in an attempt to increase the degree of universality and flexibility a motion planning has when performing under a variety of external conditions.}

\medskip
Let $p:E\to B$ be a fibration with path-connected fiber $X$. A \emph{parametrised motion planning algorithm for $p$} is a function \red{$\mathcal{A}$ assigning,} to any pair of points  $(e_1,e_2)\in E\times E$ with $p(e_1)=p(e_2)$, a continuous path $\gamma=\red{\mathcal{A}(e_1,e_2)}$ \red{in~$E$ that starts at $e_1$, ends at $e_2$, and satisfies} $p\circ\gamma=\overline{p(e_1)}$, where $\overline{b}$ \red{stands for} the constant path \red{at} $b\in B$. \red{Mathematically, $\mathcal{A}$ is a (not necessarily continuous) section of the fibration $\red{\gamma\mapsto(\gamma(0),\gamma(1))}$ defined on the fibered path space} $$E^{[0,1]}_B=\{\gamma\in E^{[0,1]}:~p\circ\gamma\text{ is a constant path}\}$$ \red{and taking values in the fibered product} $$E\times_B E=\{(e_1,e_2)\in E\times E:~p(e_1)=p(e_2)\},$$ where  $E^{[0,1]}$ stands for the free-path space on $E$. \red{In such a model, the space $B$ is meant to parametrise all possible external conditions for a given system and, for any parameter $b\in B$, the fibre $p^{-1}(b)$ represents the corresponding space of actual states of the system where motion is to be planned.}

\medskip
For practical purposes, a parametrised motion planning algorithm should depend continuously on the pair of points $(e_1,e_2)\in E\times_B E$. Indeed, if the autonomous system performs within a noisy environment, \red{then} absence of continuity could lead to instability issues in the behavior of the parametrised motion planning algorithm. In other words, continuous parametrised motion planning algorithms are robust to noise. Unfortunately, a (global) continuous parametrised motion planning algorithm for $p$ can exist only for a contractible fiber $X$ (see~\cite[Proposition 4.5]{cohen-farber-weinberger}).~Yet, if $X$ is not contractible, we could care about finding \emph{local} continuous  parametrised motion planning algorithms, i.e.,  parametrised motion planning algorithms $s$ defined only on a certain open set of $E\times_B E$, to which we refer as the domain of definition of $s$. In these terms, a \emph{ parametrised motion planner for $p$} is a set of local continuous  parametrised motion planning algorithms whose domains of definition cover $E\times_B E$. The \emph{parametrised topological complexity of~$p$}, TC$_B(X)$, is then the minimal cardinality among  parametrised motion planners for $p$, while a  parametrised motion planner for $p$ is said to be \emph{optimal} if its cardinality is TC$_B(X)$. Note that the reduced version of this invariant is presented in \cite{cohen-farber-weinberger}. Because of our application minded goals, in this work we use the unreduced version. \red{Summarizing,} the components \red{in} the parametrised motion planning problem via topological complexity are:
\begin{enumerate}
    \item The fibration $p:E\to B$ with fiber $X$. \red{Here,} a choice of a point $b\in B$ \red{in the base space} corresponds to a choice of the external conditions
for the system. 
    \item Query pairs $e=(e_1,e_2)\in E\times_B E$. The point $e_1\in E$ is the initial configuration of the query. The point $e_2\in E$ is the goal configuration.  
\end{enumerate}
In the above setting, the goal is to either describe a parametrised motion planning algorithm, i.e., describe:
\begin{enumerate}\addtocounter{enumi}{2}
\item An open covering $U=\{U_1,\ldots,U_k\}$ of $E\times_B E$.
\item For each $i\in\{1,\ldots,k\}$, a parametrised \red{motion planning algorithm,} i.e., a continuous map $s_i\colon U_i\to E^{[0,1]}_B$ satisfying $s_i(e)\left(j\right)=e_{j+1}$ for any $e=(e_1,e_2){}\in U_i$ and any $j\in\{0,1\}$,
\end{enumerate}
or, else, report that such \red{system of sections} does not exist.

\medskip
Let $X$ be a connected topological manifold of dimension at least 2. \red{Consideration of the} collision-free motion \red{planning problem} for $n$ \red{labelled} robots, each with state space $X$, in the presence of $m$ obstacles with \red{\emph{a priori}} unknown positions, \red{led Cohen, Farber and Weinberger} to study the \textit{Fadell-Neuwirth fibration}
\begin{equation}\label{fad-neu}
\pi_{n+m,m}:F(X,n+m)\to F(X,m), ~\pi_{n+m,m}(o_1,\ldots,o_m,x_1,\ldots,x_n)=(o_1,\ldots,o_m),
\end{equation}
with fiber $F(X-\{\text{ $m$ points}\},n)$, where $F(\red{Y},k)$ is the \textit{ordered configuration space}  of $k$ distinct points on $\red{Y}$ (see \cite{fadell1962configuration}). Explicitly,
\begin{equation}\label{confespa}
F(\red{Y},k)=\{(\red{y}_1,\ldots,\red{y}_k)\in \red{Y}^k\colon \red{y}_i\neq \red{y}_j\text{ for } i\neq j \},
\end{equation}
topologised as a subspace of the Cartesian power $\red{Y}^k$. \red{In such a model,} dynamics and other differential constraints \red{are ignored, focusing} primarily on the translations required to move the robot\red{s. In other words, robots and obstacles are represented by particles with} infinitesimal\red{ly small} mass \red{and volume, i.e.,} points in \red{a} Euclidean space $X=\mathbb{R}^d$. The position of the $i$-th robot is determined by $x_i\in\mathbb{R}^d$ in~(\ref{fad-neu}), while $o_j\in\mathbb{R}^d$ stands for the position of the $j$-th obstacle. In these terms, the condition $\red{y}_i\neq \red{y}_j$ in~(\ref{confespa}) reflects the collision-free and obstacle-avoidance requirements. Thus, a (local) parametrised motion planning algorithm for $\pi_{n+m,m}$ assigns to any pair of configurations $(C_1,C_2)$ in (an open set of) $F(\mathbb{R}^d,n+m)\times_{F(\mathbb{R}^d,m)} F(\mathbb{R}^d,n+m)$ a continuous curve of configurations \[\Gamma(t)\in F(\mathbb{R}^d,n+m),~~t\in [0,1],\]  such that \red{$\pi_{n+m,m}\circ\Gamma=\overline{\pi_{n+m,m}(C_1)}$ and $\Gamma\left(i\right)=C_{i+1}$ for $i\in\{0,1\}$.}

\medskip The parametrised topological complexity of $\pi_{n+m,m}$ in the case $X=\mathbb{R}^d$ has been computed by D.~Cohen, M.~Farber and S.~Weinberger in \cite{cohen-farber-weinberger} and \cite{cohen-farber.weinberger-plane}. The methods used therein are based on homotopy theory and, in particular, do not yield explicit motion planning algorithms. Inspired by our work in \cite{zapata2020multi}, we present \red{an explicit} parametrised motion planner for $\pi_{n+2,2}$ for any $d\geq 2$ even and $n\geq 1$. The planner has $2n+1$ regions of continuity and is optimal (in view of Theorem~\ref{gongra} below). The hypothesis that $d$ be even is essential for this planner. In fact, the parametrised topological complexity is one unit larger when $d$ is odd. On the other hand, the harder cases are those with $d$ even, for then the calculation of the parametrised topological complexity in \cite{cohen-farber.weinberger-plane} is based on non-constructive techniques of obstruction theory.

\medskip
\red{In Section~\ref{preliminar}} we recall well-known results about the homotopy invariance of parametrised topological complexity. In particular, in  Remark~\ref{constructing-sections-via-deformations-higher-case} we give explicit formulas describing how parametrised motion planners can be carried over from one space to another by means of a parametrised deformation. \red{This allows us to construct in Section~\ref{section2} the advertized} parametrised motion planning algorithm for  $\pi_{n+2,2}$ for any $d\geq 2$ even and $n\geq 1$. We emphasize that our \red{algorithm} work\red{s} for $m=2$, that is, \red{for} two obstacles. \red{Indeed,} the line \red{determined by the pair of obstacles} is key to our construction \red{as it} allow\red{s us} to define desingularizations $F^i$ in (\ref{desin}), sets $T_{i,j}$ in (\ref{Tij}), deformations $\varphi_i$ in (\ref{linear-transformation-obs}), parametrised homotopies $\sigma_{i,j}$ in (\ref{sigmaij}) and the  algorithm $\overline{\Gamma}$ in (\ref{algorithm-gamma}).

\medskip
The construction of optimal parametrised motion planners \red{in the presence of more obstacles is far from being obvious and apparently calls for substantial adjustments. To better appreciate the complexity of the problem, in Section~\ref{section3} we construct an optimal parametrised motion planner in the \blue{2-D} case of $\pi_{4,3}$, \blue{specifically,} we describe an algorithm for motionplanning a single point-like robot moving in \blue{$\mathbb{R}^2$} so to avoid collisions with three fixed point-like obstacles whose positions \blue{in $\mathbb{R}^2$} are \emph{a priori} unknown.}

\section{Preliminary results}\label{preliminar}

\red{After recalling from \cite{cohen-farber-weinberger, cohen-farber.weinberger-plane} the basic properties of parametrised topological complexity, we} give explicit formulas describing how parametrised motion planners can be carried over from one space to another by means of a parametrised deformation (Remark~\ref{constructing-sections-via-deformations-higher-case}).

\medskip
\red{In the setting of the previous section, consider} the evaluation fibration \begin{equation}\label{evaluation-fibration}
    \Pi:E^{[0,1]}_B\to E\times_B E,\quad\Pi(\gamma)=\left(\gamma(0),\gamma(1)\right).
\end{equation} 
A \textit{parametrised motion  planning  algorithm} is  a  section $s\colon E\times_B E\to E^{[0,1]}_B$ of  the  fibration  $\Pi$, i.e.,~ a (not necessarily continuous) map satisfying $\Pi\circ s=1_{E\times_B E}$, where $1_{E\times_B E}$ denotes the identity map. When $E\times_B E$ has the homotopy type of a CW complex, a continuous parametrised motion planning algorithm for $p$ exists if and only if the fiber $X$ is contractible (see \cite[Proposition 4.5]{cohen-farber-weinberger}), which forces the following definition. The \textit{parametrised topological complexity} TC$_B(X)$ of a fibration $p:E\to B$ with fiber $X$ is the Schwarz genus of the evaluation fibration~(\ref{evaluation-fibration}). In  other  words the parametrised topological complexity of $p$ is the smallest positive integer TC$_B(X)=k$ for which  the space $E\times_B E$ is covered by $k$ open subsets $E\times_B E=U_1\cup\cdots\cup U_k$ such that for any $i=1,2,\ldots,k$ there exists a continuous section $s_i:U_i\to E^{[0,1]}_B$ of $\Pi$ 
over $U_i$ (i.e., $\Pi\circ s_i=incl_{U_i}$, where $incl_{U_i}$ denotes the inclusion map). Thus, as noted in the introduction, we are using an unreduced notation for parametrised topological complexity.

\begin{example}{\em
Suppose that the fibers of $p:E\to B$ are convex sets. Given a pair 
of points $(e_1,e_2)\in E\times_B E$, we may move with constant velocity along the
straight line segment connecting $e_1$ and $e_{2}$. This clearly produces a continuous parametrised  algorithm
for the parametrised motion planning problem for $p$. Thus we have $\text{TC}_B(X)=1$.
}\end{example}

\red{For the trivial fibration $E=B\times F\to B$,} $\text{TC}_B(X)$ coincides with Farber`s topological complexity $\text{TC}(X)$ of the fiber $X$, which is defined in terms of motion planning algorithms for a robot moving between initial-final configurations~\cite{farber2003topological}. This  means that trivial parametrisation does not add complexity (see \cite[Example 4.2]{cohen-farber-weinberger}).

\medskip
The definition of $\text{TC}_B(X)$ deals with open subsets of $E\times_B E$ admitting continuous sections of the evaluation fibration (\ref{evaluation-fibration}). Yet, for practical purposes, the construction of explicit parametrised motion planning algorithms is usually done by partitioning the whole space $E\times_B E$ into pieces, over each of which a continuous section for~(\ref{evaluation-fibration}) is \red{given. As discussed next, under mild conditions the resulting value of the parametrised topological complexity remains unaffected.}

\medskip
Recall that a topological space $X$ is a \textit{Euclidean Neighbourhood Retract} (ENR) if it can be embedded into an Euclidean space $\mathbb{R}^d$ with an open neighbourhood $U$, $X\subset U\subset \mathbb{R}^d$, admiting a retraction $r:U\to X,$ $r\mid_U=1_X$. In addition, a subspace $X\subset \mathbb{R}^d$ is an ENR if and only if it is locally compact and locally contractible, see~\cite[Chap.~4, Sect.~8]{dold2012lectures}. This implies that finite-dimensional polyhedra, smooth manifolds and semi-algebraic sets are ENRs.

\begin{definition}
Let $E\times_B E$ be an ENR. A parametrised motion planning algorithm $s:E\times_B E\to E^{[0,1]}_B$ is said to be \textit{tame} if $E\times_B E$ splits as a pairwise disjoint union $E\times_B E=F_1\sqcup\cdots\sqcup F_k,$ where each $F_i$ is an ENR, and each restriction $s\mid_{F_i}:F_i\to E^{[0,1]}_B$ is continuous. The subsets $F_i$ in such a decomposition are called \emph{domains of continuity} for $s$.
\end{definition}

\begin{proposition}\emph{(\cite[Proposition 2.2]{rudyak2010higher})}\label{rudi}
For an ENR $E\times_B E$, $\text{\emph{TC}}_B(X)$ is the minimal number of domains of continuity $F_1,\ldots,F_k$ for tame parametrised motion planning algorithms $s:E\times_B E\to E^{[0,1]}_B$.
\end{proposition}

A tame parametrised motion planning algorithm $s:E\times_B E\to E^{[0,1]}_B$ with continuity domains $F_1,\ldots,F_k$ \red{yields an obvious motion-planning} implementation. \red{Namely,} given a pair of \red{initial-final} configurations $(C_1,C_2)\in\red{E\times_B E}$, find the subset $F_i$ such that $(C_1,C_2){}\in F_i$ and \red{take} the path $s_i(C_1,C_2)$ as output. 

\begin{remark}
 A tame parametrised motion planning  algorithm $s:E\times_B E\to E^{[0,1]}_{B}$ is called optimal when it admits $\text{TC}_B(X)$ domains of continuity. \red{At the end of} the introduction we noted that \red{the goal of} this paper \red{is the construction of} optimal parametrised motion planners. We can now be more precise: we actually construct parametrised tame motion planning algorithms with the advertized optimality property.
\end{remark}

The existence of a continuous parametrised motion planning algorithm on a subset $U$ of $E\times_B E$ implies the existence of a corresponding continuous parametrised motion planning algorithm on any subset $V$ of $E\times_B E$ deforming to $U$ within $E\times_B E$ in the \textit{parametrised context}. Such a fact is argued next in a constructive way, \red{extend}ing Example 6.4 in~\cite{farber2017configuration} \red{to} the parametrised case (the latter given for the non parametrised case). This of course suits best our implementation-oriented objectives.

\begin{remark}[Constructing parametrised motion planning algorithms via parametrised deformations]\label{constructing-sections-via-deformations-higher-case}
 Let $s_U:U\to E^{[0,1]}_B$ be a continuous parametrised motion planning algorithm defined on a subset $U$ of $E\times_B E$. Suppose a subset $V\subseteq E\times_B E$ can be continuously deformed within $E\times E$ into $V$ in the \textit{parametrised context}, i.e., there is a homotopy $H:V\times [0,1]\to E\times E$ such that $H(v,0)=v$, $H(v,1)\in U$ and $h_1(v,-),h_2(v,-)\in E^{[0,1]}_B$  for any $v\in V$, where $h_1,h_2$ denote the Cartesian components of $H$, $H=(h_1,h_2)$. As schematized in the picture
$$
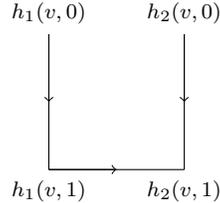
\begin{tikzpicture}[x=.6cm,y=.6cm]
\draw(0,3)--(0,0); \draw(3,3)--(3,0); 
\draw(0,0)--(3,0); 
\draw[->](0,3)--(0,1.5); 
\draw[->](3,3)--(3,1.5);
\draw[->](0,0)--(1.5,0); 
\node [below] at (0,0) {\tiny$h_1(v,1)$}; \node [above] at (0,3) {\tiny$h_1(v,0)$};
\node [below] at (3,0) {\tiny$h_2(v,1)$}; \node [above] at (3,3) {\tiny$h_2(v,0)$};
\end{tikzpicture}
$$
(where $H$ runs from top to bottom and $s_U$ runs from left to right), the path $s_U(H(v, 1))$ \red{in} $E^{[0,1]}_B$ connects in sequence the points $h_i(v,1)$, $i\in\{1,2\}$, i.e., 
$$
s_U(H(v, 1))\left(i\right)=h_{i+1}(v,1), \quad i\in\{0,1\},
$$
whereas the formula
$$
 s_V(v)(\tau) = \begin{cases}
    h_1(v,3\tau), & \hbox{$0\leq \tau\leq \frac{1}{3}$;} \\
    s_U(H(v,1))(3\tau-1), & \hbox{$\frac{1}{3}\leq \tau\leq \frac{2}{3}$;}\\
    h_2(v,3-3\tau), & \hbox{$\frac{2}{3}\leq \tau\leq 1$,}
\end{cases}.
$$
 defines a continuous section $s_V:V\to E^{[0,1]}_B$ of~(\ref{evaluation-fibration}) over $V$. Summarizing: a parametrised deformation of $V$ into $U$ and a continuous parametrised motion planning algorithm defined on $U$ determine an explicit continuous parametrised motion planning algorithm defined on $V$.
\end{remark}

The final ingredient we need is the value of TC$_{F(\mathbb{R}^d,m)}(F(\mathbb{R}^d-\{m \text{ points }\},n))$, computed by Cohen-Farber-Weinberger in \cite{cohen-farber-weinberger} and \cite{cohen-farber.weinberger-plane}.

\begin{theorem}\emph{(\cite{cohen-farber-weinberger},\cite{cohen-farber.weinberger-plane})}\label{gongra}
{For any $m\geq 2$ and $n\geq 1$, the parametrised topological complexity of the problem of collision-free motion of $n$ robots in the Euclidean $d$-space in the presence of $m$ point obstacles with unknown a priori positions is given by
\[\text{\emph{TC}}_{F(\mathbb{R}^d,m)}\left(F(\mathbb{R}^d-\{m \text{ points }\},n)\right)=\begin{cases}
2n+m,& \hbox{ if $d\geq 3$ is odd;}\\
2n+m-1,& \hbox{ if $d\geq 2$ is even}.
\end{cases}
\]}
\end{theorem}

\section{Parametrised motion planning algorithm \red{for} $\pi_{n+2,2}$ with $d\geq 2$ even}\label{section2}
We present a parametrised motion planning algorithm for $\pi_{n+2,2}$ under the assumption (in force throughout this section) that $d\geq 2$ is even. The algorithm \red{has} $2n+1$ domains of continuity.

\begin{figure}[h!]
 \centering
 \begin{tikzpicture}[x=.6cm,y=.6cm]
\draw[->](0,0)--(1,0.5); \node[below] at (0,0) {\tiny$0$}; \node[above] at (1,0.5) {\tiny$e_C$};
\draw (-6,0)--(8,7); \node[below] at (-5,0.5) {\tiny$o_1$}; \node[below] at (-2,2) {\tiny$o_2$};
\node[above] at (8,7) {\tiny$L_C$}; \draw[->](4,7)--(5,5.5);
\node[above] at (4,7) {\tiny$x_i$}; \node[below] at (5,5.5) {\tiny$p_C(x_i)$}; 
\end{tikzpicture}
\caption{The line $L_C$, its orientation $e_C$, and the projection $p_C$.}
\label{Lc}
\end{figure}

For a configuration $C=(o_1,o_2,x_1,\ldots,x_{n})\in F(\mathbb{R}^d,n+2)$, consider the affine line $L_C$ through the points $o_1$ and $o_2$, oriented in the direction of the unit vector \[e_C=\dfrac{o_2-o_1}{\mid o_2-o_1\mid},\] and let $L^\prime_C$ denote the line passing through the origin and parallel to $L_C$ (with the same orientation as $L_C$). Let $p_C:\mathbb{R}^d\to L_C$ be the orthogonal projection, and let $\overline{\text{cp}}(C)$ be the cardinality of the set $\{p_C(o_1),p_C(o_2),p_C(x_1),\ldots,p_C(x_n)\}$. Note that $\overline{\text{cp}}(C)$ ranges from $2$ to $n+2$. For $i\in\{2,\ldots,n+2\}$, let $A_i$ denote the set of all configurations $C\in F(\mathbb{R}^d,n+2)$ with $\overline{\text{cp}}(C)=i$. The various $A_i$ are ENR's satisfying
\begin{equation}\label{cerraduras2}
\overline{A_i}\subset \bigcup_{j\leq i}A_j.
\end{equation}

\subsection{Desingularization}
For a configuration $C=(o_1,o_2,x_1,\ldots,x_{n}) \in A_i$, set
\[\overline{\epsilon} (C):=\dfrac{1}{n+2}\min\{\mid p_{C}(x_r)-p_{C}(x_s)\mid\colon  p_{C}(x_r)\neq p_{C}(x_s)\}, \] here $x_i=o_i$ for $i=1,2$. In addition, for $C$ as above and $t\in[0,1]$, set 
\[F^i(C,t)=\begin{cases}(o_1,o_2,\overline{z}_1(C,t),\ldots,\overline{z}_n(C,t)), & \mbox{if $i<n+2$;}\\
 C, & \mbox{if $i=n+2$,}
 \end{cases}\]
where $\overline{z}_j(C, t)=x_j+tj\overline{\epsilon}(C)e_C$ for $j=1,\ldots,n$. This defines a continuous ``desingularization'' deformation \begin{equation}
    \label{desin} F^i:A_i\times [0,1]\to F(\mathbb{R}^d,n+2)
\end{equation} of $A_i$ into $A_{n+2}$ inside $F(\mathbb{R}^d,n+2)$ (see Figure~\ref{algorithm3}). Note that neither the lines $L_C$ and $L^\prime_ C$ nor their orientations change under the desingularization, i.e.,~$L_{F^i(C,t)}=L_C$, $L^\prime_{F^i(C,t)}=L^\prime_ C$, and $e_{F^i(C,t)}=e_C$ for all $t\in[0,1]$. Indeed, we note that $\pi_{n+2,2}(F^i(C,t))=\pi_{n+2,2}(C)$ for all $t\in[0,1]$.

\begin{figure}[h!]
 \centering
\begin{tikzpicture}[x=.4cm,y=.4cm]
\draw[->](-4,0)--(-3,0); 
\draw(-2,0)--(18,0); 
\node [below] at (-3,0) {\tiny$e_C$};
\filldraw[color=black!60, fill=black!5, very thick](-1,0) circle (0.5); \node[ ] at (-1,0) {\tiny$o_1$}; \filldraw[color=black!60, fill=black!5, very thick](16,0) circle (0.5); \node[ ] at (16,0) {\tiny$o_2$};
 \draw[->](11,0)--(12,0); \filldraw[color=black!60, fill=black!5, very thick](11,0) circle (0.5); \node[ ] at (11,0) {\tiny$1$};
 \filldraw[color=black!60, fill=black!50, very thick](12,0) circle (0.5);\node[ ] at (12,0) {\tiny$1$};
 \draw[->](4,-5)--(6,-5); \draw[->](4,-5)--(5,-5); \filldraw[color=black!60, fill=black!5, very thick](4,-5) circle (0.5); \node[ ] at (4,-5) {\tiny$2$};\filldraw[color=black!60, fill=black!50, very thick](6,-5) circle (0.5); \node[ ] at (6,-5) {\tiny$2$};
 \draw[->](4,5)--(8,5);\draw[->](4,5)--(6,5); \filldraw[color=black!60, fill=black!5, very thick](4,5) circle (0.5); \node[ ] at (4,5) {\tiny$3$};
 \filldraw[color=black!60, fill=black!50, very thick](8,5) circle (0.5); \node[ ] at (8,5) {\tiny$3$};
 \draw[dashed](4,-6)--(4,6);
\end{tikzpicture}
\caption{Desingularization.}
 \label{algorithm3}
\end{figure}

\subsection{The sets $T_{ij}$}\label{subtn322} We recall the sets $A_{ij}$ and $B_{ij}$ from \cite{zapata2020multi}. For $i,j=2,\ldots,n+2$ let 
\begin{align*}
A_{ij}:=\{(C,C^\prime)\in A_i\times A_j\colon e_C\neq -e_{C^\prime}\},\\
B_{ij}:=\{(C,C^\prime)\in A_i\times A_j\colon e_C =  -e_{C^\prime}\}.
\end{align*}
The sets $A_{ij}$ and $B_{ij}$ are ENR's (for they are semi-algebraic) covering $F(\mathbb{R}^d,n+2)\times F(\mathbb{R}^d,n+2)$ that satisfy
\begin{equation}\label{clausura}
\overline{A_{ij}}\subseteq\bigcup_{r\leq i, \;s\leq j} A_{rs}\;\;\cup \bigcup_{r\leq i, \;s\leq j} B_{rs} \quad\mbox{and}\quad \overline{B_{ij}}\subseteq \bigcup_{r\leq i,\; s\leq j} B_{rs},
\end{equation} 
in view of~(\ref{cerraduras2}). Note that each $B_{ij}$ does not intersect the subspace $F(\mathbb{R}^d,n+2)\times_{F(\mathbb{R}^d,2)} F(\mathbb{R}^d,n+2)$, i.e., \[B_{ij}\cap \left(\rule{0mm}{4mm}F(\mathbb{R}^d,n+2)\times_{F(\mathbb{R}^d,2)} F(\mathbb{R}^d,n+2)\right)=\varnothing,\] because $(C,C^\prime)\in F(\mathbb{R}^d,n+2)\times_{F(\mathbb{R}^d,2)} F(\mathbb{R}^d,n+2)$ implies that $e_C=e_{C^\prime}$ and thus $(C,C^\prime)\notin B_{ij}$. Consider subsets \begin{equation}
    \label{Tij} T_{ij}:=A_{ij}\cap \left(\rule{0mm}{4mm}F(\mathbb{R}^d,n+2)\times_{F(\mathbb{R}^d,2)} F(\mathbb{R}^d,n+2)\right).
\end{equation} The sets $T_{ij}$ are ENR's (for they are semi-algebraic) covering $F(\mathbb{R}^d,n+2)\times_{F(\mathbb{R}^d,2)} F(\mathbb{R}^d,n+2)$ that satisfy
\begin{equation}\label{clausura-rel}
\overline{T_{ij}}_{\text{rel}}\subseteq\bigcup_{r\leq i, \;s\leq j} T_{rs},
\end{equation} 
in view of~(\ref{clausura}). Here, $\overline{T_{ij}}_{\text{rel}}$ denotes the closure relative to the space $F(\mathbb{R}^d,n+2)\times_{F(\mathbb{R}^d,2)} F(\mathbb{R}^d,n+2)$, i.e., $\overline{T_{ij}}_{\text{rel}}=\overline{T_{ij}}\cap \left(F(\mathbb{R}^d,n+2)\times_{F(\mathbb{R}^d,2)} F(\mathbb{R}^d,n+2)\right)$. 

\medskip
We also consider subset $X$ of $F(\mathbb{R}^d,n+2)\times_{F(\mathbb{R}^d,2)} F(\mathbb{R}^d,n+2)$ defined by
\begin{align*}
X\hspace{.6mm}&:=\{(C,C^\prime)\in F(\mathbb{R}^d,n+2)\times_{F(\mathbb{R}^d,2)} F(\mathbb{R}^d,n+2) \colon \mbox{ with both $C$ and $C'$ colinear}\}.
\end{align*}
Here a configuration $C\in F(\mathbb{R}^d,n+2)$ is colinear if in fact $C\in F(L_C,n+2)$. 

\begin{remark}
The map $\varphi=(\varphi_1,\ldots,\varphi_{n+2}):A_{n+2}\times [0,1]\to F(\mathbb{R}^d,n+2)$ given by the formula 
\begin{equation}
    \label{linear-transformation-obs} \varphi_i(C,t)=y_i+t(p_C(y_i)-y_i),~i=1,\ldots,n+2,
\end{equation}
where $C=(o_1,o_2,x_1,\ldots,x_n)\in A_{n+2}$ and $y_1=o_1, y_2=o_2$ and $y_{i+2}=x_i$ for each $i=1,\ldots,n$, defines a continuous deformation of $A_{n+2}$ onto $F(L_C,n+2)$ inside $F(\mathbb{R}^d,n+2)$ depicted in Figure~\ref{algorithm2}. Note that, $\varphi_i(C,t)=o_i$ for any $t\in [0,1]$ and each $i=1,2$.
\end{remark}

\begin{figure}[h]
 \centering
\begin{tikzpicture}[x=.4cm,y=.4cm]
\draw[->](-4,0)--(-3,0); 
\draw(-2,0)--(18,0); 
\node [below] at (-3,0) {\tiny$e_C$};
\filldraw[color=black!60, fill=black!5, very thick](-1,0) circle (0.5); \node[ ] at (-1,0) {\tiny$o_1$}; \filldraw[color=black!60, fill=black!5, very thick](16,0) circle (0.5); \node[ ] at (16,0) {\tiny$o_2$};
 \draw[->](6,-4)--(6,0); \draw[->](6,-4)--(6,-2); \filldraw[color=black!60, fill=black!5, very thick](6,-4) circle (0.5); \node[ ] at (6,-4) {\tiny$1$}; \filldraw[color=black!60, fill=black!50, very thick](6,0) circle (0.5); \node[ ] at (6,0) {\tiny$1$};
 \draw[->](1,-5)--(1,0);\draw[->](1,-5)--(1,-2.5); \filldraw[color=black!60, fill=black!5, very thick](1,-5) circle (0.5); \node[ ] at (1,-5) {\tiny$2$}; \filldraw[color=black!60, fill=black!50, very thick](1,0) circle (0.5); \node[ ] at (1,0) {\tiny$2$};
 \filldraw[color=black!60, fill=black!5, very thick](11,0) circle (0.5);\node[ ] at (11,0) {\tiny$3$};
\end{tikzpicture}
\caption{Linear deformation of $A_{n+2}$ onto $F(L_C,n+2)$ inside $F(\mathbb{R}^d,n+2)$.}
 \label{algorithm2}
\end{figure}

\subsection{Deformations $\sigma_{ij}$} Next we define parametrised homotopies \begin{equation}
    \label{sigmaij} \sigma_{ij}:T_{ij}\times [0,1]\to F(\mathbb{R}^d,n+2)\times  F(\mathbb{R}^d,n+2),
\end{equation} deforming $T_{ij}$ into $X$, i.e.,~such that \begin{enumerate}
    \item $\sigma_{ij}((C,C^\prime),0)=(C,C^\prime)$ and $\sigma_{ij}((C,C^\prime),1)\in X$.
    \item $\pi_{n+2,2}\circ \sigma_{ij}^1((C,C^\prime),-)=\overline{\pi_{n+2,2}(C)}$ and $\pi_{n+2,2}\circ \sigma_{ij}^2((C,C^\prime),-)=\overline{\pi_{n+2,2}(C^\prime)}$, where $\sigma_{ij}^1,\sigma_{ij}^2$ denote the Cartesian components of $\sigma_{ij}$, i.e., $\sigma_{ij}=(\sigma_{ij}^1,\sigma_{ij}^2)$. Recall that $\pi_{n+2,2}(C)=\pi_{n+2,2}(C^\prime)$ for any $(C,C^\prime)\in T_{ij}$.
\end{enumerate} 

\smallskip\noindent\emph{The deformation $\sigma_{ij}$:} Given  a  pair $(C,C^\prime)\in T_{ij}$, we first apply the  desingularization  deformations $F^i(C,t)$ and $F^j(C^\prime,t)$ in order to take the pair $(C,C^\prime)$ into a pair of configurations $(C_1,C^\prime_1){}\in T_{n+2,n+2}$ (recall $\pi_{n+2,2}(C_1)=\pi_{n+2,2}(C)=\pi_{n+2,2}(C^\prime)=\pi_{n+2,2}(C_1^\prime)$). Next we apply the linear deformation~(\ref{linear-transformation-obs}), in order to take the pair $(C_1,C^\prime_1)$ into a pair of colinear configurations $(C_2,C^\prime_2){}\in X$. The deformation $\sigma_{ij}$ is the concatenation of the two deformations just described.

\subsection{Section over $X$}\label{sbsctn325}
Recall that $X\subset F(\mathbb{R}^d,n+2)\times_{F(\mathbb{R}^d,2)} F(\mathbb{R}^d,n+2)$ is the set of pairs $(C,C^\prime)$ of colinear configurations. Note that, $L_C=L_{C^\prime}=:L_{C,C'}$. We construct a continuous parametrised motion planning algorithm \begin{equation}
    \label{algorithm-gamma} \overline{\Gamma}\colon X\to F(\mathbb{R}^d,n+2)^{[0,1]}_{F(\mathbb{R}^d,2)}
\end{equation} provided $d$ is even (this is the only place where the hypothesis about the parity of $d$ is used). 

\begin{figure}[h!]
 \centering
\begin{tikzpicture}[x=.4cm,y=.4cm]
\draw[->](-4,0)--(-3,0); \draw[->](6,0)--(6,1); \draw[->](1,0)--(1,2); \draw[->](11,0)--(11,3);
\draw[->](-4,0)--(-4,1); \draw[->](6,1)--(14,1); \draw[->](1,2)--(9,2); \draw[->](11,3)--(4,3);
\draw(-2,0)--(18,0); \draw[->](14,1)--(14,0); \draw[->](9,2)--(9,0); \draw[->](4,3)--(4,0);
\node [below] at (-3,0) {\tiny$e_C$};
\node [above] at (-4,1) {\tiny$\nu(e_C)$};
\filldraw[color=black!60, fill=black!5, very thick](-1,0) circle (0.5); \node[ ] at (-1,0) {\tiny$o_1$}; \filldraw[color=black!60, fill=black!5, very thick](16,0) circle (0.5); \node[ ] at (16,0) {\tiny$o_2$};
\filldraw[color=black!60, fill=black!5, very thick](6,0) circle (0.5); \node[ ] at (6,0) {\tiny$1$}; \filldraw[color=black!60, fill=black!50, very thick](14,0) circle (0.5); \node[ ] at (14,0) {\tiny$1$};
\filldraw[color=black!60, fill=black!5, very thick](1,0) circle (0.5); \node[ ] at (1,0) {\tiny$2$}; \filldraw[color=black!60, fill=black!50, very thick](9,0) circle (0.5); \node[ ] at (9,0) {\tiny$2$};
\filldraw[color=black!60, fill=black!5, very thick](11,0) circle (0.5); \node[ ] at (11,0) {\tiny$3$}; \filldraw[color=black!60, fill=black!50, very thick](4,0) circle (0.5); \node[ ] at (4,0) {\tiny$3$};
\end{tikzpicture}
\caption{Section over $X$. Vertical arrows pointing upwards (downwards) describe the first (last) third of the path $\overline{\Gamma}^{C,C'}$, whereas horizontal arrows describe the middle third of $\overline{\Gamma}^{C,C'}$.}
 \label{algorithm1}
\end{figure}

Let $\nu$ be a fixed unitary tangent vector field on $S^{d-1}$, say $$\nu(x_1,y_1,\ldots,x_\ell,y_\ell)=(-y_1,x_1,\ldots,-y_\ell,x_\ell)$$ with $d=2\ell$. Given two configurations $C=(o_1,o_2,x_1,\ldots,x_n)$ and $C^\prime=(o_1,o_2,x^\prime_1,\ldots,x^\prime_n)$ in $F(L_{C,C^\prime},n+2)$, let $\overline{\Gamma}^{C,C'}$ be the path in the fiber $\pi_{n+2,2}^{-1}(o_1,o_2)\subset F(\mathbb{R}^d,n+2)$ from $C$ to $C^\prime$ depicted in Figure~\ref{algorithm1}. Explicitly, if $C=(o_1,o_2,x_1,\ldots,x_n)$ and $C^\prime=(o_1,o_2,x^\prime_1,\ldots,x^\prime_n)$, then the path $\overline{\Gamma}(C,C')$ in the fiber  $\pi_{n+2,2}^{-1}(o_1,o_2)\subset F(\mathbb{R}^d,n+2)$ from $C$ to $C'$ has components $(o_1,o_2,\overline{\Gamma}^{C,C'}_1,\ldots,\overline{\Gamma}^{C,C'}_n)$ defined by
$$\overline{\Gamma}^{C,C'}_{i}(t)=\begin{cases}
  \begin{array}{ll}
    x_{i}+(3ti)v(e_C), & \hbox{for $0\leq t\leq \frac{1}{3}$;} \\
    x_{i}+iv(e_C)+(3t-1)(x^\prime_{i}-x_{i}), & \hbox{for $\frac{1}{3}\leq t\leq \frac{2}{3}$;}\\
    x^\prime_{i}+i(3-3t)v(e_C), & \hbox{for $\frac{2}{3}\leq t\leq 1$.}
    \end{array}
\end{cases}$$

\subsection{Repacking regions of continuity} 
As explained in Remark \ref{constructing-sections-via-deformations-higher-case}, we can combine the continuous parametrised motion planning algorithm $\overline{\Gamma}$ with the concatenation of the parametrised deformations discussed so far to obtain continuous parametrised motion planning algorithms
\begin{equation}\label{porhoy}
T_{i,j}\to F(\mathbb{R}^d,n+2)^{[0,1]}_{F(\mathbb{R}^d,2)}, 
\end{equation}
for $i,j=2,\ldots,n+2$. The corresponding upper bound $$\text{TC}_{F(\mathbb{R}^d,2)}\left(F(\mathbb{R}^d-\{\text{ $2$ points }\},n)\right)\leq (n+1)^2$$ is improved by repacking these regions of continuity. Set
\[W_{\ell}=\bigcup_{i+j=\ell}T_{ij}\] for $\ell=4,\ldots,2n+4$. In view of (\ref{clausura-rel}), the sets assembling each $W_\ell$ are topologically disjoint in the sense that $\overline{T_{ij}}_{\text{rel}}\cap T_{i^{\prime} j^{\prime}}=\varnothing$, provided $i+j=i'+j'$ and $(i,j)\neq(i',j')$, so the sets $W_\ell$ are ENR's covering $F(\mathbb{R}^d,n+2)\times_{F(\mathbb{R}^d,2)} F(\mathbb{R}^d,n+2)$ on each of which the corresponding algorithms in~(\ref{porhoy}) assemble a continuous parametrised motion planning algorithm. We have thus constructed a tame parametrised motion planning algorithm for $\pi_{n+2,2}\colon F(\mathbb{R}^d,n+2)\to F(\mathbb{R}^d,2)$ having $2n+1$ regions of continuity $W_4,W_5,\ldots,W_{2n+4}$ (see Figure~\ref{algorithm4}).

\begin{figure}[h!]
 \centering
\begin{tikzpicture}[x=.4cm,y=.4cm]
\draw[->](-4,0)--(-3,0); 
\draw[->](-4,0)--(-4,1); 
\draw(-2,0)--(20,0); 
\node [below] at (-3,0) {\tiny$e_C$};
\node [above] at (-4,1) {\tiny$\nu(e_C)$};
\filldraw[color=black!60, fill=black!5, very thick](-1,0) circle (0.5); \node[ ] at (-1,0) {\tiny$o_1$}; \filldraw[color=black!60, fill=black!5, very thick](19,0) circle (0.5); \node[ ] at (19,0) {\tiny$o_2$};
 \draw[->](11,0)--(12,0); \draw[->](12,0)--(12,1); \draw[->](12,1)--(15,1); \draw[->](15,1)--(15,0); \draw[->](15,0)--(15,5); \draw[->](15,5)--(14,5); 
 \filldraw[color=black!60, fill=black!5, very thick](11,0) circle (0.5);\node[ ] at (11,0) {\tiny$1$};
 \filldraw[color=black!60, fill=black!50, very thick](14,5) circle (0.5);\node[ ] at (14,5) {\tiny$1$};
 \draw[dashed](14,5)--(14,0);
 \draw[->](4,-5)--(6,-5);\draw[->](6,-5)--(6,0); \draw[->](6,0)--(6,2); \draw[->](6,2)--(16,2); \draw[->](16,2)--(16,0); \draw[->](16,0)--(14,0);  
 \filldraw[color=black!60, fill=black!5, very thick](4,-5) circle (0.5); \node[ ] at (4,-5) {\tiny$2$};\filldraw[color=black!60, fill=black!50, very thick](14,0) circle (0.5); \node[ ] at (14,0) {\tiny$2$};
 \draw[->](4,5)--(8,5);\draw[->](8,5)--(8,0); \draw[->](8,0)--(8,3); \draw[->](8,3)--(17,3); \draw[->](17,3)--(17,0); \draw[->](17,0)--(17,-5); \draw[->](17,-5)--(14,-5);  
 \filldraw[color=black!60, fill=black!5, very thick](4,5) circle (0.5); \node[ ] at (4,5) {\tiny$3$};
 \filldraw[color=black!60, fill=black!50, very thick](14,-5) circle (0.5); \node[ ] at (14,-5) {\tiny$3$};
 \draw[dashed](4,-4.5)--(4,4.5);
 \draw[dashed](14,-0.5)--(14,-4.5);
\end{tikzpicture}
\caption{The motion planning algorithm for $\pi_{n+2,2}$.}
 \label{algorithm4}
\end{figure}

\section{Parametrised motion planning algorithm \red{for} $\pi_{4,3}$ with \blue{$d=2$}}\label{section3}
We present a parametrised motion planning algorithm for $\pi_{4,3}$ \blue{in the 2-D case}
The algorithm \red{has} four domains of continuity, \red{so its optimality follows from Theorem~\ref{gongra}. As in} Section~\ref{section2}, we consider \red{a} unitary tangent vector field  $\nu$ on $S^{\blue{1}}$, \red{say the one} given by \blue{$\nu(x_1,y_1)=(-y_1,x_1)$.}

\smallskip
For a configuration $C=(o_1,o_2,o_3,x)\in F(\mathbb{R}^{\blue{2}},4)$, consider the affine line $L_C$ through the points $o_1$ and $o_2$ oriented in the direction of the unit vector \[e_C=\dfrac{o_2-o_1}{\mid o_2-o_1\mid},\] and let $L^\perp_{C}$ denote the \red{affine} line \blue{perpendicular to $L_C$} that passes through the point $o_1$ \blue{and is} oriented in the direction of the unit vector $\nu(e_C)$.
$$
\begin{tikzpicture}[x=.5cm,y=.5cm]
\draw[dashed, ->](0,0)--(4,0);
\draw[dashed, ->](1,-1)--(1,2);
\node at (1,0) {$\bullet$};
\node at (3,0) {$\bullet$};
\node[below] at (.6,-.2) {$o_1$};
\node[below] at (3,-.2) {$o_2$};
\node at (4.7,0) {$L_C$};
\node at (0,1.8) {$L^\perp_C$};
\end{tikzpicture}
$$
Let \red{$p_C:\mathbb{R}^{\blue{2}}\to L_C$ and $p^\perp_C:\mathbb{R}^{\blue{2}}\to L^\perp_C$} be the orthogonal projections, and \red{set}
\begin{eqnarray*}
\text{cp}^{\red{\perp}}_o(C) &=&\mid \{p^\perp_C(o_1),p^\perp_C(o_2),p^\perp_C(o_3)\}\mid,\\
\text{cp}^{\red{\perp}}(C) &=& \mid \{p^\perp_C(o_1),p^\perp_C(o_2),p^\perp_C(o_3),p^\perp_C(x)\}\mid,\\
\text{cp}_o(C) &=&\mid \{p_C(o_1),p_C(o_2),p_C(o_3)\}\mid,\\
\text{cp}(C) &=& \mid \{p_C(o_1),p_C(o_2),p_C(o_3),p_C(x)\}\mid,
\end{eqnarray*} where $\mid S\mid$ denotes the cardinality of the set $S$.
Note that \red{$$\text{cp}^{\red{\perp}}_o(C)\in\{1,2\},\;\;\text{cp}^\perp(C)\in\{1,2,3\},\;\;\text{cp}_o(C)\in\{2,3\},\;\;\text{cp}(C)\in\{2,3,4\},$$ although not all combinations are achievable for a point $(C,C')$ in the fibered product $F(\mathbb{R}^{\blue{2}},4)\times_{F(\mathbb{R}^{\blue{2}},3)} F(\mathbb{R}^{\blue{2}},4)$. To be precise, for $i,j\in\{1,2\}$, $r,s\in \{2,3\}$ and $k,l\in\{3,4\}$, consider the subsets $T^{1}_{i,j}$, $T^{2,2}_{r,s}$ and $T^{2,3}_{k,l}$ of $F(\mathbb{R}^{\blue{2}},4)\times_{F(\mathbb{R}^{\blue{2}},3)} F(\mathbb{R}^{\blue{2}},4)$ consisting of the pairs $(C,C')$ satisfying the following list of conditions:} 
\begin{align*}
\text{In \,} T^{1}_{i,j} \colon\hspace{1mm}
\;&\text{cp}^{\red{\perp}}_o(C)=1,~\text{cp}^{\red{\perp}}(C)=i \mbox{\hspace{2mm}and \,} \text{cp}^{\red{\perp}}(C')=j.\\
\text{In \,} T^{2,2}_{r,s} \colon
\;&\text{cp}^{\red{\perp}}_o(C)=2,~\text{cp}_o(C)=2,~\text{cp}(C)=r~\text{and \,cp}(C')=s.\\
\text{In \,} T^{2,3}_{k,l}\colon
\;&\text{cp}^{\red{\perp}}_o(C)=2,~\text{cp}_o(C)=3,~\text{cp}(C)=k~\text{and \,cp}(C')=l.
\end{align*}
\red{Thus, for $(C,C')\in T^{1}_{i,j}$, the three common obstacles in $C$ and $C'$ lie in \blue{$L_C$,} whereas the non-obstacle in $C$ (respectively $C'$) lies in \blue{$L_C$} if and only if $i=1$ (respectively, $j=1$). Likewise, for $(C,C')\in T^{2,2}_{r,s}$, the four relative positions of the three common obstacles in $C$ and $C'$ can be depicted as} 
$$
\begin{tikzpicture}[x=.4cm,y=.4cm]
\draw[dashed, ->](0,0)--(4,0);
\draw[dashed, ->](1,-2.5)--(1,2.5);
\draw[dashed, ->](3,-2.5)--(3,2.5);
\node at (1,0) {$\bullet$};
\node at (3,0) {$\bullet$};
\node[below] at (.5,-.05) {$o_1$};
\node[below] at (3.5,-.05) {$o_2$};
\node at (4.7,0) {$L_C$};
\node[left] at (.6,2.8) {$L^\perp_C$};
\node[right] at (3.4,2.8) {$\overline{L}^\perp_C$};
\node at (3,1.5) {$\circ$};\node[right] at (3,1.5) {$o_3$};
\node at (1,1.5) {$\circ$};\node[left] at (1,1.5) {$o_3$};
\node at (1,-1.7) {$\circ$};\node[left] at (1,-1.7) {$o_3$};
\node at (3,-1.7) {$\circ$};\node[right] at (3,-1.7) {$o_3$};
\end{tikzpicture}
$$
\red{whereas the non-obstacle in $C$ ($C'$, respectively) lies on $$L^\perp_C\cup\overline{L}^\perp_C$$ if and only if $r=2$ ($s=2$, respectively). Lastly, for $(C,C')\in T^{2,3}_{k,l}$, the third common obstacle $o_3$ lies outside $$L_C\cup L^\perp_C\cup\overline{L}^\perp_C,$$ while the non-obstacle in $C$ ($C'$, respectively) determines a fourth projection on $L_C$ if and only if $k=4$ ($l=4$, respectively). We thus have:}

\begin{corollary}\label{lapart}
The various \red{ENR's} $T^1_{i,j}$, $T^{2,2}_{r,s}$ and $T^{2,3}_{k,l}$ \red{give a partition of the fibered product} $F(\mathbb{R}^{\blue{2}},4)\times_{F(\mathbb{R}^{\blue{2}},3)} F(\mathbb{R}^{\blue{2}},4)$.
\end{corollary}

\red{Continuous parametrized motion planning algorithms on the various $T^{*,*}_{*,*}$ are described next. In each case, motion is meant to be performed at constant speed along the indicated path. The following conventions are in force in the next pictures: \emph{(i)} The obstacle $o_3$ is sometimes omitted when it lies in $L_C$ and is not relevant. \emph{(ii)} The auxiliary dashed oriented lines $L_C$ and $L^\perp_C$ are drawn without specifying their names. \emph{(iii)} The positive (negative, respectively) hemiplane $H_{C,+}$ ($H_{C,-}$, respectively) determined by $L_C$ is the one located in the positive (negative, respectively) $L^\perp_C$-direction, likewise we have positive and negative hemiplanes $H^\perp_{C,+}$ and $H^\perp_{C,-}$ determined by $L^\perp_C$, where signs are determined by the $L_C$-direction. \emph{(iv)} We set $C=(o_1,o_2,o_3,x)$ and $C'=(o_1,o_2,o_3,x')$, and let $d(u,v)$ stand for the Euclidean distance between the points $u,v\in\mathbb{R}^{\blue{2}}$.}

\begin{figure}[h!]
$$
\begin{tikzpicture}[x=.35cm,y=.35cm]
\draw[thin,dashed,->](0,0)--(9,0);
\draw[thin,dashed,->](1,-1.5)--(1,2);
\node at (1,0) {$\bullet$};
\node at (5,0) {$\bullet$};
\node[below] at (.5,-.2) {$o_1$};
\node[below] at (5,-.2) {$o_2$};
\node at (7,0) {$\bullet$};\node[below] at (7,0) {$x'$};
\node at (3,0) {$\bullet$};\node[below] at (3,-.2) {$x$};
\draw[thick,->](3,0)--(3,1.5);
\draw[thick,->](3,1.5)--(7,1.5);
\draw[thick,->](7,1.5)--(7,.3);
\draw[dotted,->](8,1.5)--(8,0);
\draw[dotted,->](8,0)--(8,1.5);
\node at (8.5,.8) {$1$};
\draw[thin,dashed,->](14+0,0)--(14+9,0);
\draw[thin,dashed,->](28+0,0)--(28+9,0);
\draw[thin,dashed,->](14+1,-1.5)--(14+1,2);
\draw[thin,dashed,->](28+1,-1.5)--(28+1,2);
\node at (14+1,0) {$\bullet$};
\node at (28+1,0) {$\bullet$};
\node at (14+5,0) {$\bullet$};
\node at (28+5,0) {$\bullet$};
\node[below] at (14+.5,-.2) {$o_1$};
\node[below] at (28+.5,-.2) {$o_1$};
\node[below] at (14+5,-.2) {$o_2$};
\node[below] at (28+5,-.2) {$o_2$};
\node at (14+8,1) {$\bullet$};\node[right] at (14+8,1) {$x'$};
\node at (28+8,-1.5) {$\bullet$};\node[right] at (28+8.1,-1.5) {$x'$};
\node at (14+2.5,-3) {$\bullet$};\node[below] at (14+2.5,-3) {$x$};
\node at (28+2.5,-3) {$\bullet$};\node[below] at (28+2.5,-3) {$x$};
\node[left] at (14+5,1.5) {$o_3$};\node at (14+5,1.5) {$\bullet$};
\node[right] at (28+6,1.5) {$o_3$};\node at (28+6,1.5) {$\bullet$};
\draw[thick,->](14+2.5,-3)--(14+2.5,3);
\draw[thick,->](28+2.5,-3)--(28+2.5,3);
\draw[thick,->](14+2.5,3)--(22,3);
\draw[thick,->](28+2.5,3)--(22+14,3);
\draw[thick,->](22,3)--(22,1.31);
\draw[thick,->](22+14,3)--(14+22,-1.2);
\draw[dotted,->](8,1.5)--(8,0);
\draw[dotted,->](8,0)--(8,1.5);
\node at (8.5,.8) {$1$};
\draw[dotted,->](12+8,1.5+1.5)--(12+8,0+1.5);
\draw[dotted,->](12+8,0+1.5)--(12+8,1.5+1.5);
\node at (12+8.5,.8+1.5) {$1$};
\draw[dotted,->](13+12+8,1.5+1.5)--(13+12+8,0+1.5);
\draw[dotted,->](13+12+8,0+1.5)--(13+12+8,1.5+1.5);
\node at (13+12+7.5,.8+1.5) {$1$};
\end{tikzpicture}
$$
\caption{$T^{1}_{1,1}$ (left), $T^{2,2}_{3,3}$ (center) and $T^{2,3}_{4,4}$ (right) }
\label{T111}
\end{figure}

\red{Parametrised motion planning in $T^1_{1,1}$, $T^{2,2}_{3,3}$ and $T^{2,3}_{4,4}$ uses the paths in $H_{C,+}$ depicted in Figure~\ref{T111} and constructed in terms of three lines, namely, the $L^\perp_C$-parallel lines through $x$ and $x'$, and the $L_C$-parallel line having $p_C$ projection $1+\max\{p_C(o_1),p_C(o_2),p_C(o_3)\}$.}

\begin{figure}
$$
\begin{tikzpicture}[x=.35cm,y=.35cm]
\draw[thin,dashed,->](0,0)--(9,0);
\draw[thin,dashed,->](1,-1.5)--(1,2);
\node at (1,0) {$\bullet$};
\node at (5,0) {$\bullet$};
\node[below] at (.5,-.2) {$o_1$};
\node[below] at (5,-.2) {$o_2$};
\node at (7,1.5) {$\bullet$};\node[right] at (7,1.5) {$x'$};
\node at (3,0) {$\bullet$};\node[below] at (3,-.2) {$x$};
\draw[thick,->](3,0)--(6.84,1.47);
\draw[thin,dashed,->](15+0,0)--(15+9,0);
\draw[thin,dashed,->](16,-1.5)--(16,2);
\node at (16,0) {$\bullet$};
\node at (20,0) {$\bullet$};
\node[below] at (15.5,-.2) {$o_1$};
\node[below] at (20,-.2) {$o_2$};
\node at (15+7,0) {$\bullet$};\node[below] at (15+7,0) {$x'$};
\node at (18,1.5) {$\bullet$};\node[above] at (18,1.6) {$x$};
\draw[thick,->](18,1.5)--(15+6.85,0.1);
\end{tikzpicture}
$$
\caption{$T^1_{1,2}$ (left) and $T^1_{2,1}$ (right)}
\label{Tstraight}
\end{figure}

\begin{figure}
$$
\begin{tikzpicture}[x=.35cm,y=.35cm]
\draw[thin,dashed,->](0,0)--(9,0);
\draw[thin,dashed,->](1,-2)--(1,2);
\node at (1,0) {$\bullet$};
\node at (5,0) {$\bullet$};
\node[below] at (.4,-.2) {$o_1$};
\node[below] at (5,-.2) {$o_3$};
\node at (6.5,0) {$\bullet$};
\node[below] at (6.5,-.2) {$o_2$};
\node at (7,-2) {$\bullet$};\node[right] at (7,-2) {$x'$};
\node at (8,1.5) {$\bullet$};\node[above] at (8,1.6) {$x$};
\draw[thick,->](8,1.5)--(3,1.5);
\draw[thick,->](3,1.5)--(3,-2);
\draw[thick,->](3,-2)--(6.825,-2);
\draw[dotted,->](1,2.3)--(3,2.3);
\draw[dotted,->](3,2.3)--(1,2.3);
\node[above] at (2,2.3) {$\delta_1$};
\draw[thin,dashed,->](15,0)--(15+9,0);
\draw[thin,dashed,->](16,-2)--(16,3);
\node at (16,0) {$\bullet$};
\node at (20,0) {$\bullet$};
\node[below] at (15.1,.2) {$o_1$};
\node[below] at (20.6,-.2) {$o_2$};
\draw[thin,dashed,->](20,-2)--(20,3);
\node at (16,1.5) {$\bullet$};
\node[left] at (16,1.5) {$o_3$};
\node at (16,-1.5) {$\bullet$};\node[left] at (16,-1.5) {$x'$};
\node at (20,1.5) {$\bullet$};\node[right] at (20,1.4) {$x$};
\draw[thick,->](20,1.5)--(18,1.5);
\draw[thick,->](18,1.5)--(18,-1.5);
\draw[thick,->](18,-1.5)--(16.15,-1.5);
\draw[dotted,->](16,2.3)--(18,2.3);
\draw[dotted,->](18,2.3)--(16,2.3);
\node[above] at (17,2.3) {$\delta_2$};
\end{tikzpicture}
$$
\caption{$T^1_{2,2}$ (left) and $T^{2,2}_{2,2}$ (right)}
\label{porenmedio}
\end{figure}

\red{As depicted in Figure~\ref{Tstraight}, parametrised motion planning in $T^1_{1,2}$ and $T^1_{2,1}$ uses straight lines. Parametrised motion planning in $T^1_{2,2}$ and $T^{2,2}_{2,2}$ uses the paths depicted in Figure~\ref{porenmedio} and constructed in terms of the $L_C$-parallel lines through $x$ and $x'$ and the $L^\perp_C$-parallel line in $H^\perp_{C,+}$ whose distance to $o_1$ is $\delta_1=\frac12\min\{d(o_1,o_2),d(o_1,o_3)\}$, in the case of $T^1_{2,2}$, and $\delta_2=\frac12d(o_1,o_2)$, in the case of $T^{2,2}_{2,2}$.}

\begin{figure}[h!]
$$
\begin{tikzpicture}[x=.35cm,y=.35cm]
\draw[thin,dashed,->](0,0)--(9,0);
\draw[very thin,dashed,->](0,1)--(9,1);
\draw[thin,dashed,->](1,-2)--(1,5);
\node at (1,0) {$\bullet$};
\node at (5,0) {$\bullet$};
\node[below] at (.2,0) {$o_1$};
\node[below] at (5.8,0) {$o_2$};
\draw[thin,dashed,->](5,-2)--(5,3);
\node at (1,2) {$\bullet$};
\node[left] at (1,2) {$o_3$};
\node at (1,4) {$\bullet$};
\node[left] at (1,4.55) {$x$};
\draw[thick,->](1,4)--(3,4);
\draw[thick,->](3,4)--(3,1);
\draw[very thin,dashed,->](0,4)--(9,4);
\draw[thick,->](3,1)--(4,1);
\draw[thick,->](4,1)--(4,-2);
\node[below] at (4,-2) {$x'$};
\node at (4,-2.3) {$\bullet$};
\draw[thin,dashed,->](15,0)--(15+9,0);
\draw[very thin,dashed,->](15+0,1)--(15+9,1);
\draw[thin,dashed,->](15+1,-2)--(15+1,5);
\node at (15+1,0) {$\bullet$};
\node at (15+5,0) {$\bullet$};
\node[below] at (15+.2,0) {$o_1$};
\node[below] at (15+5.8,0) {$o_2$};
\draw[thin,dashed,->](15+5,-2)--(15+5,3);
\node at (15+1,2) {$\bullet$};
\node[left] at (15+1,2) {$o_3$};
\node at (15+1,4) {$\bullet$};
\node[left] at (15+1,4.75) {$x'$};
\draw[thick,->](15+3,4)--(15+1.2,4);
\draw[thick,->](15+3,1)--(15+3,4);
\draw[very thin,dashed,->](15+0,4)--(15+9,4);
\draw[thick,->](15+4,1)--(15+3,1);
\draw[thick,->](15+4,-2)--(15+4,1);
\node[below] at (15+4,-2.4) {$x$};
\node at (15+4,-2.3) {$\bullet$};
\end{tikzpicture}
$$
\caption{Parametrised motion planning in $T^{2,2}_{2,3}$ (left) and $T^{2,2}_{3,2}$ (right)}
\label{t2223}
\end{figure}

\red{Parametrised motion planning in $T^{2,2}_{2,3}$ and $T^{2,2}_{3,2}$ uses the paths depicted in Figure~\ref{t2223} and constructed in terms of four lines $\ell_1,\ldots,\ell_4$. For instance, in the case of $T^{2,2}_{2,3}$, $\ell_1$ is the $L_C$-parallel line through $x$; $\ell_2$ is the $L^\perp_C$-parallel line through the middle point between $o_1$ and $o_2$; $\ell_3$ is the $L_C$-parallel line through the middle point between $o_3$ and the obstacle $o_i$ having $i\in\{1,2\}$ and $p_C(o_3)=p_C(o_i)$; $\ell_4$ is the $L^\perp_C$-parallel line through $x'$.}

\smallskip
\red{Parametrised motion planning in $T^{2,3}_{3,3}$ is best pictured in terms of the grid in Figure~\ref{grid}, where vertical (horizontal, respectively) dashed lines represent the three (two, respectively) different values in $\{p_C(o_1),p_C(o_2),p_C(o_3)\}$ ($\{p^\perp_C(o_1),p^\perp_C(o_2),p^\perp_C(o_3)\}$, respectively) determined by an element in $T^{2,3}_{3,3}$. Solid lines are then constructed to be right in between two consecutive dashed lines, except for the right-most vertical solid line that is chosen to be one unit to the right of the right-most dashed vertical line. In such a setting, obstacles are located at the intersections of dashed lines (there are only six possibilities), whereas $x$ and $x'$ are located along vertical dashed lines. Parametrised motion planning from $x$ to $x'$ then uses the simple path constructed in terms of the three solid lines in Figure~\ref{grid} together with the $L_C$-parallel lines connecting $x$ and $x'$ to the first solid vertical line on their right.}

\begin{figure}[h!]
$$
\begin{tikzpicture}[x=.35cm,y=.35cm]
\draw[very thin,dashed](2,2)--(16,2);
\draw[very thick](2,4)--(16,4);
\draw[thin,dashed](2,6)--(16,6);
\draw[very thick](14,0)--(14,8);
\draw[very thick](6,0)--(6,8);
\draw[very thick](10,0)--(10,8);
\draw[very thin, dashed](4,0)--(4,8);
\draw[very thin, dashed](8,0)--(8,8);
\draw[very thin, dashed](12,0)--(12,8);
\end{tikzpicture}
$$
\caption{Grid for $T^{2,3}_{3,3}$}
\label{grid}
\end{figure}

\red{Parametrised motion planning in $T^{2,3}_{3,4}$ and $T^{2,3}_{4,3}$ uses the strategy in the previous paragraph, with a single modification. Namely, in the case of $T^{2,3}_{3,4}$ ($T^{2,3}_{4,3}$, respectively), so that $x'$ ($x$, respectively) does not lie on some of the vertical dashed lines of Figure~\ref{grid}, the corresponding $L_C$-parallel line through $x'$ ($x$, respectively) is replaced by the $L^\perp_C$-parallel line connecting $x'$ ($x$, respectively) to the solid lines in Figure~\ref{grid}.}

\medskip
\red{The discussion above is still not enough to get the desired optimal parametrised motion planner with 4 domains. We need a suitable repacking of the various $T$'s. Explicitly, we consider the partition of} $F(\mathbb{R}^{\blue{2}},4)\times_{F(\mathbb{R}^{\blue{2}},3)} F(\mathbb{R}^{\blue{2}},4)$ \red{given by} the ENR's 
\begin{eqnarray}
W_1 &=& T^{1}_{1,1}\cup T^{2,2}_{3,3}\cup T^{2,3}_{4,4},\nonumber\\
W_2 &=& T^{1}_{2,2}\cup T^{2,2}_{2,2},\label{w2}\\
W_3 &=& T^{1}_{1,2}\cup T^{1}_{2,1}\cup T^{2,2}_{2,3}\cup T^{2,2}_{3,2}\cup T^{2,3}_{3,3} \text{\;\;\;\red{and}}\label{w3}\\
W_4 &=& T^{2,3}_{3,4}\cup T^{2,3}_{4,3}.\label{w4}
\end{eqnarray}

\begin{proposition}
\red{The parametrised motion planning algorithms on the various $T$'s assemble a parametrised motion planner with domains of definition $W_i$ for $1\leq i\leq4$.}
\end{proposition}
\begin{proof}
\red{Continuity of the parametrised motion planning algorithm for $W_1$ follows by direct inspection of Figure~\ref{T111}. We prove continuity in the other three cases by observing that the unions in (\ref{w2})--(\ref{w4}) are topological. In $W_2$ we have $$\overline{T^1_{2,2}}\cap T^{2,2}_{2,2}=\varnothing$$ because the condition $\text{cp}^\perp_o(C)=1$ defining $T^1_{2,2}$, which is inherited by $\overline{T^1_{2,2}}$, is incompatible with the defining condition $\text{cp}^{\red{\perp}}_o(C)=2$ in $T^{2,2}_{2,2}$. Likewise, the equality $$T^1_{2,2}\cap\overline{T^{2,2}_{2,2}}=\varnothing$$ holds since the condition $\text{cp}_o(C)=2$ defining $T^{2,2}_{2,2}$, which is inherited by $\overline{T^{2,2}_{2,2}}$, is incompatible with the condition $\text{cp}_o(C)=3$ forced in $T^1_{2,2}$. On the other hand, the equalities $$\overline{T^{2,3}_{3,4}}\cap T^{2,3}_{4,3}=\varnothing=T^{2,3}_{3,4}\cap\overline{T^{2,3}_{4,3}}$$ in $W_4$ follow by looking at conditions $\text{cp}(C)$ and $\text{cp}(C')$, respectively. Finally, the fact that the first four $T$-pieces of $W_3$ are topologically separated from the rest of the pieces comes by looking at:
\begin{itemize}
\item $\text{cp}^\perp(C)$ for the $T^1_{1,2}$ piece;
\item $\text{cp}^\perp(C')$ for the $T^1_{2,1}$ piece;
\item $\text{cp}(C)$ for the $T^{2,2}_{2,3}$ piece;
\item $\text{cp}(C')$ for the $T^{2,2}_{3,2}$ piece.
\end{itemize}
The topologically-separated condition for the last piece $T^{2,3}_{3,3}$ of $W_3$ is a bit more ellaborated:
\begin{itemize}
\item $\overline{T^{2,3}_{3,3}}\cap T^1_{1,2}=\varnothing$ because of the respective conditions on $\text{cp}(C)$;
\item $\overline{T^{2,3}_{3,3}}\cap T^1_{2,1}=\varnothing$ because of the respective conditions on $\text{cp}(C')$;
\item $\overline{T^{2,3}_{3,3}}\cap T^{2,2}_{2,3}=\varnothing$ as $p_C(x')\in\{p_C(o_1),p_C(o_2),p_C(o_3)\}$ holds in $\overline{T^{2,3}_{3,3}}$ but not in $T^{2,2}_{2,3}$;
\item $\overline{T^{2,3}_{3,3}}\cap T^{2,2}_{3,2}=\varnothing$ as $p_C(x)\in\{p_C(o_1),p_C(o_2),p_C(o_3)\}$ holds in $\overline{T^{2,3}_{3,3}}$ but not in $T^{2,2}_{3,2}$.
\end{itemize}}
\end{proof}

\bibliographystyle{plain}

\end{document}